\documentclass{article}
\pdfoutput=1
\usepackage[preprint]{neurips_2021}
\usepackage[utf8]{inputenc} 
\usepackage[T1]{fontenc}    
\usepackage{hyperref}       
\usepackage{url}            
\usepackage{booktabs}       
\usepackage{amsfonts}       
\usepackage{nicefrac}       
\usepackage{microtype}      
\usepackage{xcolor}         
\usepackage{graphicx}
\usepackage{multirow}
\usepackage{siunitx}
\usepackage{float}
\usepackage{wrapfig}
\usepackage{booktabs}
\usepackage{caption}
\usepackage{amsmath}
\usepackage{adjustbox}
\usepackage{authblk}

\title{Adaptive Multi-Resolution Attention with Linear Complexity}

\newcommand*\samethanks[1][\value{footnote}]{\footnotemark[#1]}
\author{Yao Zhang\thanks{these authors contributed equally to this work}, \textsuperscript{\rm 1}
	\qquad Yunpu Ma\samethanks, \textsuperscript{\rm 1}
	\qquad Thomas Seidl, \textsuperscript{\rm 1}
	\qquad Volker Tresp \textsuperscript{\rm 1,2}\\
    \textsuperscript{\rm 1} Institute of Informatics, LMU Munich, 
    \textsuperscript{\rm 2} Corporate Technology, Siemens AG \\
   	Yao.Zhang@campus.lmu.de, \qquad yunpu.ma@dbs.ifi.lmu.de\\
   	seidl@dbs.ifi.lmu.de, \qquad volker.tresp@siemens.com
}

\begin{document}

\maketitle

\begin{abstract}
\label{abstract}
Transformers have improved the state-of-the-art across numerous tasks in sequence modeling. Besides the quadratic computational and memory complexity w.r.t the sequence length, the self-attention mechanism only processes information at the same scale, i.e., all attention heads are in the same resolution, resulting in the limited power of the Transformer. To remedy this, we propose a novel and efficient structure named \textit{\textbf{Ada}ptive \textbf{M}ulti-\textbf{R}esolution \textbf{A}ttention} (\textit{AdaMRA} for short), which scales linearly to sequence length in terms of time and space. 
Specifically, we leverage a multi-resolution multi-head attention mechanism, enabling attention heads to capture long-range contextual information in a coarse-to-fine fashion. Moreover, to capture the potential relations between query representation and clues of different attention granularities, we leave the decision of which resolution of attention to use to query, which further improves the model's capacity compared to vanilla Transformer. In an effort to reduce complexity, we adopt kernel attention without degrading the performance. 
Extensive experiments on several benchmarks demonstrate the effectiveness and efficiency of our model by achieving state-of-the-art \textbf{speed-memory-accuracy trade-off}.
To facilitate \textit{AdaMRA} utilization by the scientific community, the code implementation will be made publicly available. 
  
\end{abstract}

\section{Introduction}
\label{introduction}
The recent emergence of the Transformer has drastically reshaped the landscape of natural language processing research. Transformers have demonstrated superior performance in a wide variety of tasks, such as machine translation \citep{vaswani2017attention}, natural language inference \citep{williams2017broad}, text classification \citep{howard2018universal}, question answering \citep{rajpurkar2016squad}, automatic speech recognition \citep{dong2018speech}, image generation \citep{parmar2018image} and image captioning \citep{xu2015show}. The key innovation in Transformers is the introduction of a multi-head self-attention mechanism, which models pairwise interaction of the input sequence, regardless of their distance from each other. This operation has been shown quite effective.

Nonetheless, despite several notable successes of Transformers, computing the attention matrix, which is their key component, also turns out to be a major efficiency bottleneck due to its quadratic time and space complexity with respect to the sequence length. Therefore, the maximum sequence length is restricted by the amount of memory available. This inherent limitation of Transformers has prevented them from being successfully applied to domains requiring longer sequence lengths, like document classification. Further, building large Transformer-based models in practice is notoriously expensive. Although the fine-tuning stage is relatively inexpensive, the memory issue still restricts the scenarios in which these models can be used. Besides the computational cost, qualitative analysis  

\begin{wrapfigure}{r}{0.55\textwidth}
\label{intro_fig}
\vspace{-10pt}
  \begin{center}
    \includegraphics[width=0.55\textwidth]{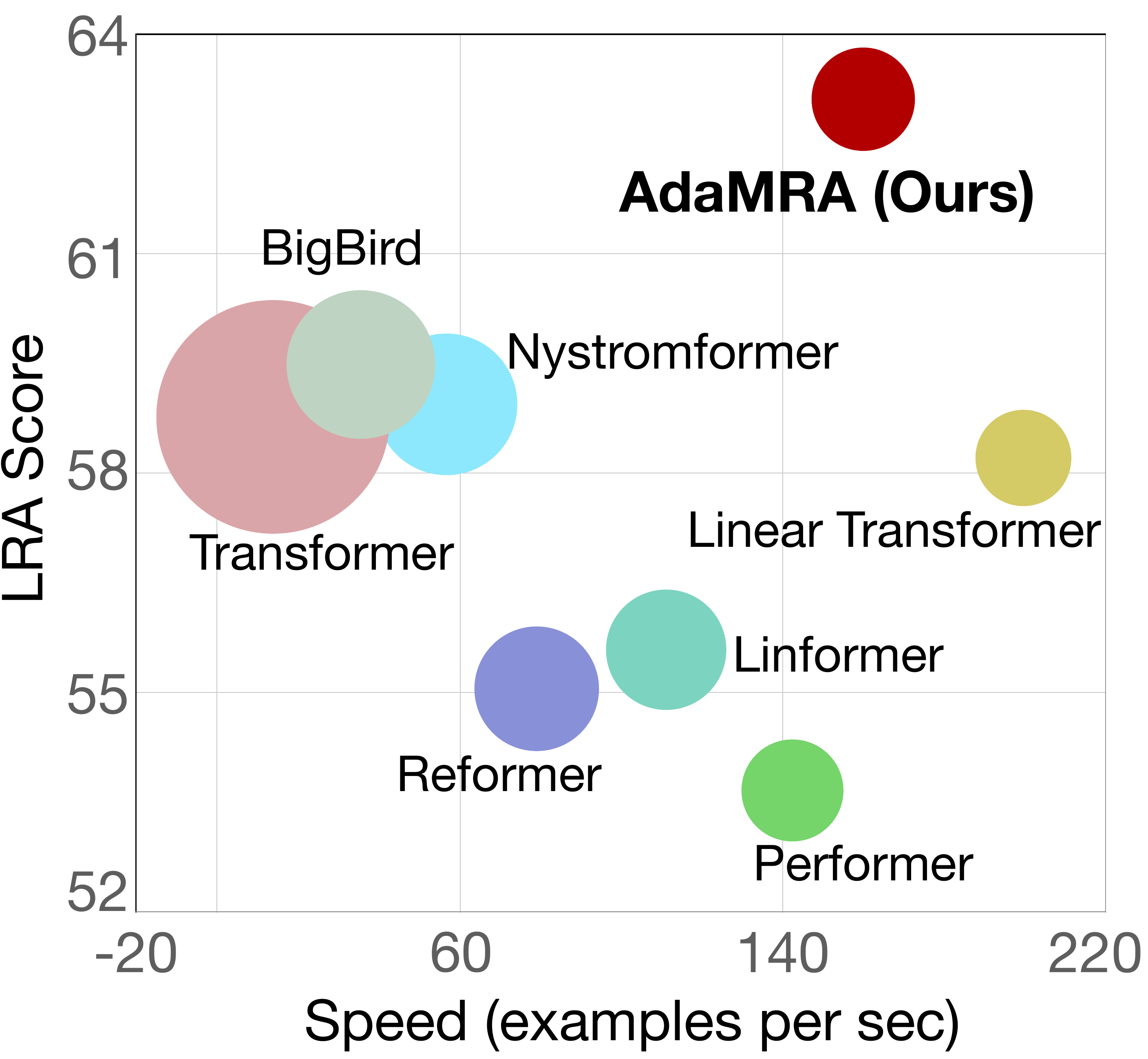}
  \end{center}
\vspace{-10pt}
  \caption{Trade-off between accuracy (LRA score), model speed (examples per sec) and memory footprint (size of circle). Our \textit{AdaMRA} achieves state-of-the-art speed-memory-accuracy trade-off.}
\end{wrapfigure}

of attention heads \citep{vaswani2017attention} suggests that heads tend to favor flatter or more peaked distributions depending on what phenomena they capture. Thus, using an extremely long sequence may limit the power of the model. 

To this end, a wide spectrum of efficient, fast Transformers has been proposed to tackle these limitations. For instance, \citep{beltagy2020longformer,wang2020transformer,tay2020sparse,child2019generating,zaheer2020big,correia2019adaptively,qiu2019blockwise,roy2021efficient,kitaev2020reformer,zhao2019explicit,vyas2020fast} addresses the problematic complexity by limiting the amount of keys that each query attends to. However, these methods either break long-term dependency or hurt the time efficiency. There is also a long line of research on using dense attention matrix but defined by low-rank kernels substituting softmax \citep{katharopoulos2020transformers,choromanski2020rethinking,xiong2021nystr,peng2021random}. Although these approaches have achieved better speed-memory-accuracy trade-off, they still suffer from the aforementioned limitations of the self-attention mechanism. Another prominent line of work is to increase the memory capacity \citep{sukhbaatar2019adaptive,ye2019bp,rae2019compressive}. However, these works still process information at the same scale. 

Besides the computational cost, the self-attention mechanism of all models mentioned above only processes information at the same scale, i.e., all attention heads are in the same resolution. However, inspired by the fact that the information in most of the domain has a hierarchical structure, for instance, word- and sentence-level information in the domain of text/language, low- and high-level features in the domain of image, etc., we suggest processing information in a coarse-to-fine fashion could be beneficial for capturing hierarchically structured information. We thus propose \textit{\textbf{Ada}ptive \textbf{M}ulti-\textbf{R}esolution \textbf{A}ttention} (\textit{AdaMRA}), a linear time and space attention that captures long-distance dependencies in a coarse-to-fine manner. To be more precise, unlike vanilla Transformer, which maintains a constant resolution throughout all attention heads, \textit{AdaMRA} employs multi-resolution attention heads that vary in the level of abstraction. Moreover, to capture the potential relations between query representation and clues of different attention granularities, each query is routed to the corresponding attention head. Furthermore, we adopt kernel attention \citep{katharopoulos2020transformers} without sacrificing the performance.

We evaluate the proposed method on the Long-Range-Arena (LRA) benchmark \citep{tay2020long} and show that \textit{AdaMRA} achieves promising performance while having a linear computational complexity with respect to the sequence length. Impressively, as shown in Figure 1, the average LRA scores increased by 4.32 and 3.66 from vanilla Transformer and the previous best performing model BigBird, respectively. In terms of time and space efficiency, \textit{AdaMRA} is around 10 times faster than vanilla Transformer on GPU, while 5 times smaller in GPU running memory occupation. 

\section{Related Work}
\label{related}
In this section, we briefly review the most relevant works that aim to address the Transformers' large memory and computational requirements. 

\paragraph{Efficient Self-Attention}
A conceptual way of reducing the complexity of the full attention is to limit the number of accessible elements to the attention. \citep{qiu2019blockwise,zaheer2020big,beltagy2020longformer,child2019generating} achieve this by using fixed, predefined patterns such as local windows and block patterns of fixed stride. Another line of work here is to consider which part of the inputs should be attended to by learning to assign tokens to buckets or clusters before performing attention. \citep{kitaev2020reformer} uses locality sensitive hashing to group together token, \citep{roy2021efficient,vyas2020fast} employs online $k$-means to learn the space-partitioning centroids, and \citep{tay2020sparse} sort keys in a block-wise fashion. However, they may lack the flexibility to look at the full sequence and thus restrict the model capacity to capture long-distance dependencies. Moreover, additional computation steps required by some approaches (e.g., LSH in \citep{kitaev2020reformer}) might undermine their final efficiency gains. Unlike these works, our method uniquely incorporates pooling-based compression to capture the context information of different scales with only a small additional computation budget while maintaining excellent performance.

\paragraph{Kernel Attention} 
Another method is to improve efficiency by leveraging low-rank approximations of the softmax attention matrix. Katharopoulos et al. \citep{katharopoulos2020transformers} interprets the $Softmax$ as a kernel and approximate the attention matrix via kernel approximation. Subsequently, this strategy is also employed by \citep{xiong2021nystr,peng2021random,choromanski2020rethinking}. In \citep{xiong2021nystr}, the approximation of standard softmax attention is based on adapting the Nystr\"om method, while \citep{peng2021random,choromanski2020rethinking} leverage random feature methods to approximate the softmax function. Although these approaches have achieved better speed-memory-accuracy trade-off, the performance of these methods is still affected by the quality of approximation and the fully-connected nature of self-attention in Transformer, which, as suggested by \citep{guo2019star}, is not a good inductive bias. 

\paragraph{Increasing Memory Capacity}
Memory is crucial for many tasks. However, extending the memory span is computationally expensive due to the attention mechanism's quadratic time and space complexity. Several recent works have proposed strategies to increase the memory capacity of Transformers. BP-Transformer \citep{ye2019bp} is designed to incorporate the common-sense inductive bias of the hierarchical linguistic structure within the sentence, i.e., each query attends to context information from fine-grain to coarse-grain as the relative distance increase. \citep{rae2019compressive} uses some pooling operator (e.g., max/mean pooling) to reduce the number of memories in the past, where all memories are equally compressed regardless of the content of the current query. In \citep{sukhbaatar2019adaptive}, each attention head separately learns its temporal context size from data. The works mentioned above focus on increasing memory capacity without actually changing the memory resolution. Our work differs from theirs in that we focus on capturing long-term dependencies in a multi-resolution fashion, which, in turn, indirectly reduces our model's memory footprint.

\section{Model}
In this section, we formalize the proposed method. In Section \ref{revisit}, we first briefly revisit the attention mechanism and present its computational complexity. We then introduce \textit{AdaMRA} in Section 3.2. An interpretation to \textit{AdaMRA} is provided in Section \ref{inter}. We close by practically analyzing \textit{AdaMRA}'s complexity in Section \ref{complexity}.

\label{AMR}
\subsection{Revisiting Self-Attention and its Linearization}
\label{revisit}
The self-attention function calculates, for every token, a weighted average of the feature representations of all other tokens with a weight proportional to a normalized similarity score between representations. 
Formally, let $X=\{x^{(1)},...,x^{(n)}\}\in \mathbb{R}^{n\times d}$ denotes an input sequence comprising $n$ tokens of dimension $d$.
Given three matrices $Q,K$ and $V$, which is linear projections of the layer's input $X$,
\begin{equation}
\centering
	Q=XW^Q,\   K=XW^K, \  V=XW^V,
\end{equation}
where $Q,K,V\in \mathbb{R}^{n\times d}$ and  $W^Q,W^K,W^V\in \mathbb{R}^{d\times d}$. Following common terminology, $Q$, $K$, and $V$ are referred to as the \textit{queries}, \textit{keys}, and \textit{values}, respectively. 
The \textit{keys} are used to compute a similarity score between each item and \textit{query}. Then, weight the \textit{values} of each item at each query context using the normalized similarity score.  
The attention outputs the weighted sum of the values by the similarity score between the queries and keys. Thus, the generalized attention function for any similarity function can be written as: 
\begin{equation}
\label{eq_sim}
	Attention(Q_i,K,V)=Score(Q_i,K,V)
	=\frac{\sum^n_{j=1}sim(Q_i,K_j)V_j}{\sum^n_{j=1}sim(Q_i,K_j)}
\end{equation}
According to \citep{vaswani2017attention}, the unified similarity function can take the form of $Softmax$. Therefore, the quadratic complexity emerges from the computation of the similarity score between every pair of tokens.

In order to define an attention function, $sim(\cdot)$ in Eq. \ref{eq_sim} needs to be a non-negative function, which includes all kernels $k(x,y):\mathbb{R}^{F} \times \mathbb{R}^{F} \to \mathbb{R}_+$ \citep{katharopoulos2020transformers}. Given such a kernel with a feature representation $\phi(x)$, we can rewrite the generalized attention function (Eq. \ref{eq_sim}) as follows,
\begin{equation}
\begin{aligned}
\label{eq_linear}
Attention(Q_i,K,V)
=	\frac{\sum^n_{j=1}\phi(Q_i)^T\phi(K_j)V_j}{\sum^n_{j=1}\phi(Q_i)^T\phi(K_j)}=	\frac{\phi(Q_i)^T\sum^n_{j=1}\phi(K_j)V_j^T}{\phi(Q_i)^T\sum^n_{j=1}\phi(K_j)}
\end{aligned}	
\end{equation}
$\sum^n_{j=1}\phi(K_j)V_j^T$ and $\sum^n_{j=1}\phi(K_j)$ could be reused for every query, therefore reducing the complexity from quadratic to linear both in terms of memory and computation. 

Considering the fact that different parts of a sequence may be relevant in different ways, multi-head attention was introduced in Transformer. Assuming there are $H$ attention heads, this is simply the application of Eq. \ref{eq_sim} in parallel $H$ times, each with a different, learned linear transformation that allows specialization:
\begin{equation}
	MultiHead(Q,K,V)=Concat(Head_1,...,Head_H)W^O
\end{equation}
\begin{equation}
	Head_h(Q,K,V)=Attention(QW_h^Q,KW_h^K,VW_h^V)
\end{equation}
where $W_h^Q, W_h^K\in \mathbb{R}^{d\times d_k}$, $W_h^V\in \mathbb{R}^{d\times d_v}$ are the matrices that project the \textit{queries}, \textit{keys}, and \textit{values} into the $h$-th subspace, respectively; $W^O\in \mathbb{R}^{Hd_v\times d}$ is the matrix that computes a linear transformation of the heads, with, typically, $Hd_v=Hd_k=d$.

\subsection{Adaptive Multi-Resolution Attention}
\label{AMRA}

\begin{figure}
\label{model_fig}
  \centering
    \includegraphics[width=1\textwidth]{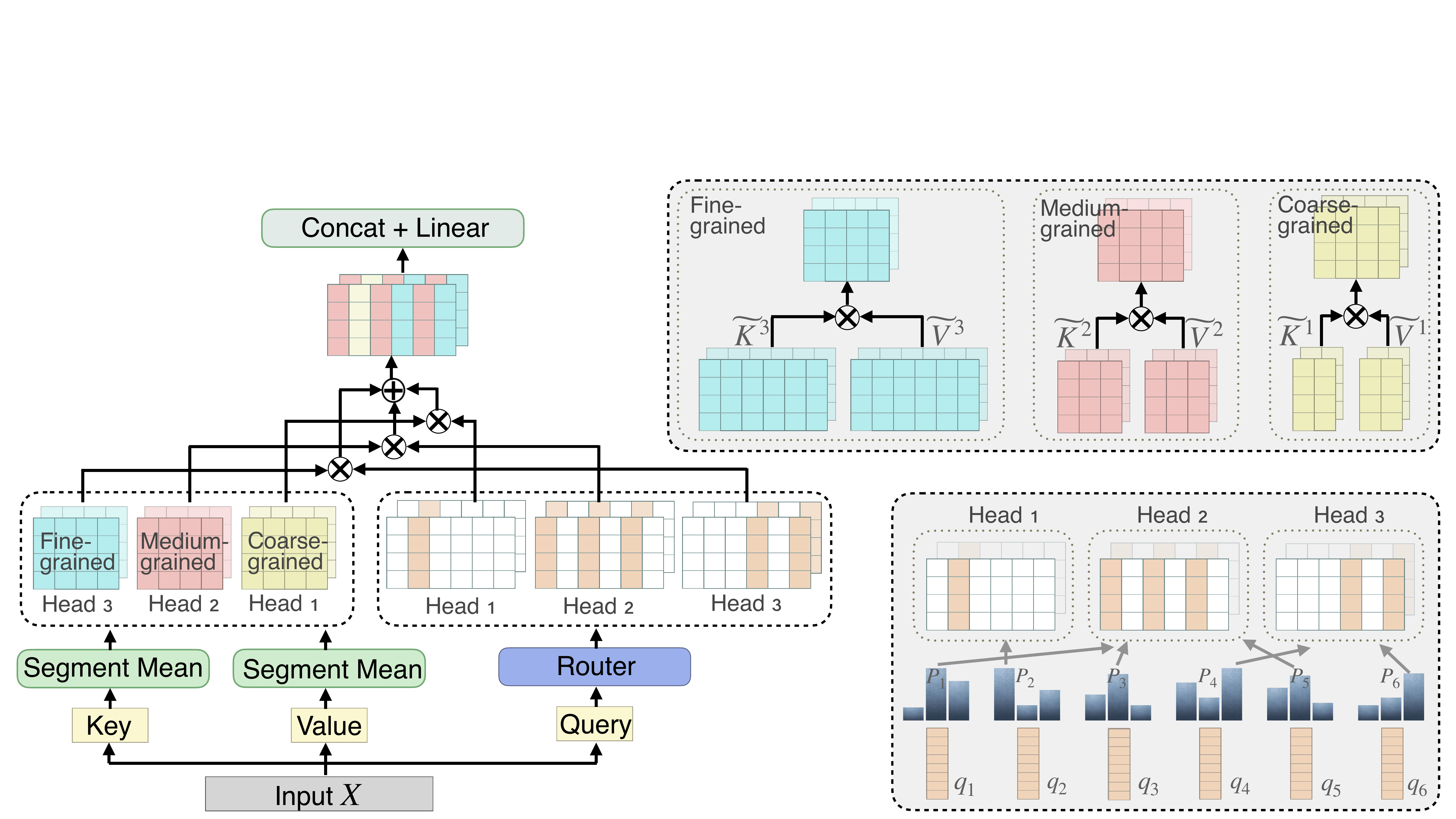}
  \caption{The architecture of \textit{AdaMRA}.  
  In this example, the number of heads $H=3$, the number of subheads $S=2$, sequence length $n=6$ and compression rate $c=(\nicefrac{1}{1},\nicefrac{1}{2},\nicefrac{1}{3})$. Top right shows the construction of multi-resolution memory/context. Bottom right diagrams 6 tokens being routed.}
\end{figure} 
To capture hierarchically structured information effectively, we propose \textit{AdaMRA}. The main idea is to employ the multi-resolution attention heads in a coarse-to-fine fashion and enable the query to choose between different resolutions of attention. This process is done independently for each layer, allowing queries in different layers to attend to contexts of different resolutions. We describe \textit{AdaMRA} in the context of a single Transformer layer and omit the layer index for brevity. 

In \textit{AdaMRA}, the input sequence $X\in \mathbb{R}^{n\times d}$ still pass through three linear layers to form the \textit{queries} $Q\in \mathbb{R}^{n\times d}$, \textit{keys} $K\in \mathbb{R}^{n\times d}$, and \textit{values} $V\in \mathbb{R}^{n\times d}$, where $n$ is sequence length and $d$ is the embedding dimension. For each attention head $h$, we define a compression rate $c_{h}$, a higher value indicates more fine-grained compressed information. 
To encode the context information, we produce compressed \textit{keys} and \textit{values} using certain compressive operations, which can be selected from $k$-means clustering \citep{vyas2020fast,roy2021efficient}, projection \citep{wang2020linformer}, and convolution \citep{rae2019compressive}, etc. For the sake of computation efficiency, we employ segment means \citep{xiong2021nystr} to  compress the original $(n\times d)$-dimensional $K$ and $V$ into $(m_h\times d)$-dimensional compressed $\widetilde{K}^{h}$ and $\widetilde{V}^{h}$, where $h$ denotes $h$-th head and $m_h=nc_{h}$ is the number of landmarks of head $h$. To be more precise, given the compression rate $c_{h}$ for head $h$, we separate the $n$ \textit{keys}/\textit{values} into $m_h$ segments. In our experiments, $n$ is divisible by $m$. If this is not the case in practice, we can pad inputs to a length divisible to $m$. Note that to obtain multi-resolution attention, compression rate of each head is different. Such compression strategy can not only guarantee the preservation of important information, but also simplify the model since $c$ is usually a small number. More details regarding the impact of compression rate is provided in Section \ref{experiment}. 

With compressed \textit{key} $\widetilde{K}^{h}$ and \textit{value} $\widetilde{V}^{h}$ available in hand, attention head are now in different resolution. To capture the potential relations between \textit{query} representation and clues of different attention granularities, we let \textit{query} itself choose which resolution of attention to use, which is conditioned on the information encoded in \textit{query}'s representation. This is accomplished by adding a Router \citep{shazeer2017outrageously} before the attention layer (see Figure 2). The router takes a token representation $q_i$ as an input and then routes this to the best-determined expert, i.e., attention head. Specifically, we adopt $F(\cdot)$, a parameterized function, for projecting \textit{query} $q_i$ from $d$ dimensions to $H$ dimensions, where $H$ is the number of heads. We then normalize this value via a $Softmax$. Each \textit{query} is routed to the head with the highest router probability $P$. In practice, we mask out the tokens that are not routed to the current head. Formally,
\begin{equation}
	P=Softmax(F(Q)),\ \ \ \ \ \  F(Q)=QW,
\end{equation}
where $F(\cdot)$ is a parameterized function, $Q\in \mathbb{R}^{n\times d}$, the learnable parameter $W\in \mathbb{R}^{d\times H}$, and the router probability $P\in \mathbb{R}^{n\times H}$. 

Finally, in pursuit of efficiency, we adopt kernel attention \citep{katharopoulos2020transformers} while calculating attention using Eq. \ref{eq_linear},
\begin{equation}
\begin{aligned}
Attention(Q_i^{h},\widetilde{K}^{h},\widetilde{V}^{h})
	=\frac{\sum^N_{j=1}sim(Q_i^{h},\widetilde{K}_j^{h})\widetilde{V}_j^{h}}{\sum^N_{j=1}sim(Q_i^{h},\widetilde{K}_j^{h})}
	=\frac{\phi(Q_i^{h})^T\sum^n_{j=1}\phi(\widetilde{K}_j^{h})(\widetilde{V}_j^{h})^T}{\phi(Q_i^{h})^T\sum^n_{j=1}\phi(\widetilde{K}_j^{h})},
\end{aligned}
\end{equation}
where $Q_i^{h}$ is $i$-th \textit{query} that is routed to $h$-th head, $\widetilde{K}^{h}$ and $\widetilde{V}^{h}$ are the compressed \textit{query} and \textit{value} of $h$-th head. For our experiments, we employ $ReLU$ as the feature function $\phi$ (see Section 4.3 for feature function analysis). 

As suggested by recent works on interpreting attention head roles, separate attention heads may learn to look for various relationships between tokens \citep{voita2019analyzing}. Thus, in practice, we use the same strategy to split the head into multiple subheads, whose resolution is the same as the original head, allowing the model to jointly attend to information at different positions from different representation subspaces. For our experiments, all attention heads have same number of subheads.
Multi-head \textit{AdaMRA} is thus defined as: 
\begin{equation}
AdaMRA(Q,K,V)=(\sum^H_{h=1}Head_h)W^O,\   
Head_h=Concat(subhead_{h_1},...,subhead_{h_S}),
\end{equation}
where $Q,K,V\in\mathbb{R}^{n\times d}$, $W^O\in\mathbb{R}^{Sd_v\times d}$ is learned matrices, $d_v=d/S$ is the hidden dimension of the projection subspace, $H$ is the number of heads, $S$ is the number of subheads, and $h_S$ denotes the $S$-th subhead of $h$-th head. The $s$-th subhead of $h$-th head is defined as:
\begin{equation}
	subhead_{h_s}=Attention(Q^{h}W_{h_s}^Q,\widetilde{K}^{h}W_{h_s}^K,\widetilde{V}^{h}W_{h_s}^V),
\end{equation}
where $W_{h_s}^Q,W_{h_s}^K\in\mathbb{R}^{d\times d_k}$, $W_{h_s}^V\in\mathbb{R}^{d\times d_v}$ are the matrices that project the \textit{queries}, \textit{keys} and \textit{values} into the $h_s$-th subspace, respectively. For our experiments, we set $Sd_v=Sd_k=d$.

\subsection{Interpretation of AdaMRA}
\label{inter}
Intuitively, one can think of $\phi(\widetilde{K}^{h})^T\widetilde{V}^{h}$ as a global description/memory of the input sequence that the \textit{query} will perform attention over. As discovered by previous works, global and multi-scale representations are useful. Therefore, to combine the low-level details and high-level semantics, each attention head has different memory scales, corresponding to a different semantic aspect of the entire input. For instance, coarse memory and fine-scale memory could correspond to the summary of paragraph and the word representation, respectively. To further enhance feature expression ability, we leave the decision of which resolution of attention to use to \textit{query}. Thus, a \textit{query} can choose between the different resolutions of memory based on its own representation with more flexibility.

\subsection{Efficiency Advantage}
\label{complexity}
We now show the efficiency advantage of \textit{AdaMRA} in memory and computation. Assuming we have $H$ head with $m_h$ landmarks each, the landmark selection using segment means takes $O(n)$, where $n$ is sequence length. The usage of kernel attention \citep{katharopoulos2020transformers} eliminates the $O(n^2)$ terms from both the memory and computational complexities of the module. Instead, the computation of global description/memory ($\phi(\widetilde{K}^{h})^T\widetilde{V}^{h}$) of dimensionality $d$ and new values take $O(m_hd^2)$ and $O(nd^2)$, respectively. 
Consequently, the total cost of \textit{AdaMRA} scales as $O(Hn+\sum^H_{h=1}m_hd^2+Hnd^2)$, i.e., scales linearly with respect to the sequence length $n$. In the following section we will show that  a small $\sum^H_{h=1}m_h$ (typically smaller than $n$) is enough for achieving good performance, which further increase the efficiency advantage of \textit{AdaMRA} over vanilla Transformer. 

\section{Experiments}
\label{experiment}
In this section, we validate the \textit{AdaMRA} in terms of computational cost, memory consumption, and accuracy on long-range context tasks in the LRA benchmark \citep{tay2020long} and show that \textit{AdaMRA} performs consistently better than baselines, suggesting that leveraging multi-resolution attention head is reasonable and effective. The experimental results show the superior ability of \textit{AdaMRA} in modeling the long-range context.  

\subsection{Experiment Settings}
LRA is a suite of five general and challenging tasks designed to evaluate how well Transformers capture long-term dependencies from different modalities such as text, natural and synthetic images, and mathematical expressions requiring similarity, structural and visual-spatial reasoning. For a complete description of the objectives and datasets, we refer the reader to \citep{tay2020long}.

\label{lra}

\paragraph{Tasks} Tasks used for comparison are as follows: (1) \texttt{Long Listops}, designed to investigate the model's capability of reasoning hierarchically structured data in a long-context scenario. We use a version of the ListOps dataset \citep{nangia2018listops} of sequence lengths of up to 2K. (2)
\texttt{Byte-Level Text Classification} aims to test the models' ability to deal with compositionality as it is required to compose characters into words into higher-level phrases. We use the IMDb reviews dataset \citep{maas2011learning} of a fixed max length of 4K. (3) 
\texttt{Byte-Level Document Retrieval} uses the AAN dataset \citep{radev2013acl} to investigate a model's ability to encode and store compressed representation. The model learns a similarity score between two documents. Each document has a sequence length of 4K. (4) \texttt{Image Classification} This task serves as a test of how well models are able to capture the 2D spatial relations between input pixels. We use the CIFAR-10 dataset \citep{krizhevsky2009learning} for this task. (5) \texttt{Pathfinder} In this task, we are interested in the model's ability to capture long-range spatial dependencies. The model makes a binary decision on whether two points are connected by a path \citep{linsley2018learning}.

\paragraph{Baselines}
We base our evaluation on six recently proposed efficient Transformer models. Aside from the vanilla Transformer \citep{vaswani2017attention}, we compare our model against other efficient self-attention variants, including Reformer \citep{kitaev2020reformer}, Linear Transformer \citep{katharopoulos2020transformers}, Performer \citep{choromanski2020rethinking}, Linformer \citep{wang2020linformer}, Big Bird \citep{zaheer2020big} and Nystr\"omformer\citep{xiong2021nystr}.

\paragraph{Implementation Details}
To ensure fair comparisons for all models, we train a two-layer Transformer. The embedding dimension is 64, and the hidden dimension is 128. Our data split, preprocessing, and training procedure follow those of \citep{xiong2021nystr}. For all models, we use the default PyTorch implementation.  

\subsection{Performance Comparison}

\paragraph{Accuracy Comparison}

\begin{table}
 \caption{Experimental results on LRA benchmark. We report accuracy (higher is better) of different models. The best model is in boldface, and the second-best is underlined. Transformer's, Reformer's, Linformer's, Performer's and Nystr\"omformer's numbers are due to \citep{xiong2021nystr}. Asides from Linear Transformer, we achieve consistent results reported in \citep{tay2020long}. The implementation of Linear Transformer is based on the official published code. Avg: average accuracy across all tasks. \textit{AdaMRA} significantly outperforms other Transformer models among all tasks, with +4.32, +3.66 in average accuracy against vanilla Transformer and previous best performing model BigBird, respectively.}
  \vspace{10pt}
  \centering
  \begin{tabular}{c|ccccc|c}
    \toprule
    Model & ListOps	& Text	& Retrieval	& Image	& Pathfinder & Avg \\
    \midrule
    Transformer& 37.10 & 65.02 & 79.35 & 38.20 & 74.16 & 58.77     \\
    \midrule
    BigBird    	&\underline{38.55} &63.90 &\underline{81.50} &38.30 & \underline{74.89}     &\underline{59.43}\\
    Reformer& 19.05 & 64.88 & 78.64 & \underline{43.29} & 69.36 & 55.04      \\
    Linformer& 37.25 & 55.91 & 79.37 & 37.84 & 67.60 & 55.59      \\
    Linear Trans.&37.35&64.15&81.10 & 38.20& 70.20 &58.20   \\
    Performer& 18.80 & 63.81 & 78.62 & 37.07 & 69.87 & 53.63      \\
    Nystr\"omformer& 37.15 & \underline{65.52} & 79.56 & 41.58 & 70.94 & 58.95     \\
    \midrule
	Ours &\textbf{40.40} &\textbf{68.44} & \textbf{84.83}&\textbf{46.00} & \textbf{75.77}&  \textbf{63.09}  \\
    \bottomrule
  \end{tabular}
 \end{table}

We compare Transformers with respect to their performance. Table 1 reports LRA score of several Transformer models, including BigBird (the previous best performing model in terms of LRA score \citep{tay2020long}), variants with linear complexity (Linformer, Linear Transformer, Performer and Nystr\"omformer), and Reformer. 

Our model brings consistently considerable performance boosts over the baseline models to all tasks. Specifically, our model achieves an average score of 63.09 on the LRA benchmark, increasing 4.32, 4.89, and 9.46 absolute points from vanilla Transformer, Linear Transformer, and Performer. Besides, our model also achieves significant performance gain compared to the previous best performing model BigBird. This might be attributed to the fact that, in contrast to BigBird, our attention layer is able to exchange information globally on the entire sequence. 

In addition, it is worth mentioning that our model boosts the score by 3.42 and 5.48 on the tasks \texttt{Text} (n=4K) and \texttt{Retrieval} (n=4K) compare with vanilla Transformer, suggesting that our model is advantageous in tasks that require large sequence length.
More importantly, we notice the performance gains, especially on the \texttt{Image Classification} task, outperforming vanilla Transformer by 7.8, which indicates that the inductive bias of \textit{AdaMRA} plays a substantial role in this task. Thus, our model has the capability to capture 2D spatial relations.

\paragraph{Speed and Memory Comparison}
\begin{table}
  \caption{Comparison of training time and peak memory consumption on various input sequence lengths. Efficiency improvements in comparison with the vanilla Transformer are in brackets. The best model is in boldface, and the second-best is underlined. Performer-32 denotes Performer self-attention module using a feature map of 32 dimensions. Nystr\"omformer-32 denotes Nystr\"omformer self-attention module using 32 landmarks. Ours-($c_h$) denotes \textit{AdaMRA} self-attention module with four attention heads using a compression rate of $c_h$ on each head. \textit{AdaMRA} offers favorable memory and time efficiency over standard self-attention and is almost as fast and compact as kernel-based Transformer (Linear Transformer and Performer).}
   \vspace{10pt}
  \centering
  \scalebox{0.9}{
  \begin{tabular}{ccccccc}
    \toprule
    \multirow{2}{*}{Model}&
    \multicolumn{3}{c}{Running time (ms)} &
    \multicolumn{3}{c}{Peak Memory Usage (MB)}  \\
    \cmidrule(lr){2-4} 
    \cmidrule(lr){5-7} 
         & 1K & 2K & 4K & 1K & 2K & 4K  \\
    \midrule
    Transformer &82(1$\times$)&272(1$\times$)&1007(1$\times$)&1713(1$\times$)&3829(1$\times$)&10327(1$\times$) \\
    \midrule
    BigBird &104(0.79$\times$)&211(1.3$\times$)&420(2.4$\times$)&1815(0.9$\times$)&2835(1.4$\times$)&4921(2.1$\times$) \\
    
	Reformer &53(1.5$\times$)&103(2.6$\times$)&200(5.0$\times$)&1467(1.2$\times$)&2007(1.9$\times$)&3125(3.3$\times$) \\
	
	Linformer &37(2.2$\times$)&71(3.8$\times$)&139(7.2$\times$)&1499(1.1$\times$)&1978(1.9$\times$)&3035(3.4$\times$) \\
	Linear Transformer &\textbf{22(3.7$\times$)}&\textbf{38(7.1$\times$)}&\textbf{69(14.6$\times$)}&\textbf{1113(1.5$\times$)}&\textbf{1353(2.8$\times$)}&\textbf{1741(5.9$\times$)} \\
    Performer-32 &\underline{30(2.7$\times$)}&54(5.0$\times$)&\underline{100(10.7$\times$)}&\textbf{1171(1.5$\times$)}&\underline{1439(2.7$\times$)}&\underline{1917(5.4$\times$)} \\
    Nystr\"omformer-32 &76(1.1$\times$)&160(1.7$\times$)&233(4.3$\times$)&2627(0.6$\times$)&3381(1.1$\times$)&4685(2.2$\times$)\\
    \midrule
Ours-(\nicefrac{1}{2},\nicefrac{1}{4},\nicefrac{1}{8},\nicefrac{1}{16}) &33(2.4$\times$)&54(5.0$\times$)&98(10.3$\times$)&\underline{1239(1.4$\times$)}&1581(2.4$\times$)&2201(4.7$\times$) \\
Ours-(\nicefrac{1}{4},\nicefrac{1}{8},\nicefrac{1}{16},\nicefrac{1}{32}) &32(2.6$\times$)&\underline{53(5.1$\times$)}&96(10.5$\times$)&\underline{1201(1.4$\times$)}&1565(2.4$\times$)&2117(4.9$\times$) \\
    \bottomrule
  \end{tabular}}
  \end{table}

To better illustrate the boosted efficiency, we compare Transformers with respect to their computational and memory requirements. We use the IMDb dataset with a batch size of 16 for all runs. 
Table 2 shows peak allocated GPU memory and required time of the sequence lengths $\{1K,2K,4K\}$ for several efficient Transformer models. We benchmark all models' speed and memory on a Tesla T4 GPU with 16GB of memory. All compared models are of the same size as those described above.

The overall fastest models are kernel-based models (Performer and Linear Transformer). The model with the smallest memory footprint is the Linear Transformer, coming in at 1741 MB compared to 10327 MB for the vanilla Transformer at 4K.
Our method comes in a close second and is almost as fast as the fastest one. Similar to speed, our model is also relatively compact and is almost as compact as Linear Transformer and Performer. 
Importantly, our model speeds up over the vanilla Transformer by about 10.4$\times$ on 4K sequence length and requires only about 20\% of the memory of the vanilla Transformer at 4K. As sequence length increases, the training time speed-up and memory savings are even more dramatic.

Notably, kernel-based models are fast and compact at the cost of relatively lower quantitative performance (see Table 1). In contrast, our model is competitive in both accuracy and efficiency, as Figure 1 shows. Besides, our analysis indicates that \textit{AdaMRA} efficiency gains are especially notable on long sequences, suggesting that \textit{AdaMRA} will be particularly useful in tasks that require large sequence length, fast training speed, or low memory footprints.

\paragraph{Speed-Memory-Accuracy Tradeoff Comparison}

\begin{table}
 \caption{Comparison of SMAT score (higher is better). Normalized values are in brackets. Speed stands for examples per second. The best model is in boldface, and the second-best is underlined. \textit{AdaMRA} significantly outperforms other Transformer models.}
  \vspace{10pt}
  \scalebox{0.95}{ 
  \centering
  \begin{tabular}{c|ccc|c}
    \toprule
    Model & Speed & Peak Memory Usage (MB)& LRA Score	& SMAT Score	 \\
    \midrule
    Transformer&14.5  (0.00)&6645 (1.00)&58.77 (0.54)&0.54     \\
    \midrule
    BigBird&36.4  (0.12)&2917 (0.30)&\underline{59.43 (0.61)}&1.43\\
    Reformer&80.0   (0.35)&2023 (0.13)&55.04 (0.15)&1.37      \\
    Linformer&111.1 (0.52)&2003 (0.12)&55.59 (0.21)& 1.61     \\
    Linear Transformer&\textbf{200.0 (1.00)}&\textbf{1353 (0.00)}&58.20 (0.48)&\underline{2.48}   \\
    Performer-32&142.9 (0.69)&\underline{1439 (0.02)}&53.63 (0.00)&   1.67   \\
    Nystr\"omformer-32&57.1    (0.23)&2687 (0.25)&58.95 (0.56)& 1.54\\
  	\midrule
	Ours &\underline{160.0  (0.78)}&\underline{1475 (0.02)}&\textbf{63.09 (1.00)}& \textbf{2.76} \\
    \bottomrule
  \end{tabular}}
 \end{table}
 
 In real-world scenarios, speed, memory and accuracy are three important aspects of performance. When analyzed separately, these performance variables sometimes lead to contradictory conclusions. To avoid such conflicts, we integrate all three aspects into a single measure, speed-memory-accuracy tradeoff (SMAT), which is defined as,  
 \begin{equation}
 	SMAT=S_{norm}+(1-M_{norm})+Acc_{norm}
 \end{equation}
 where $S_{norm}$, $M_{norm}$, $Acc_{norm}$ are normalized speed (examples per sec), peak memory usage as well as LRA score after applying the  MinMaxScaler. In this experiment, we use the IMDb dataset of the sequence length 4K with a batch size of 8 for all runs. As shown in Table 3, AdaMRA consistently outperforms other variants by a large margin in terms of SMAT score, indicating that our AdaMRA achieves state-of-the-art speed-memory-accuracy trade-off.
 
\subsection{Ablation Study}
\label{ablation}
\paragraph{Compression Rate}
\begin{table}
  \caption{Ablation studies of \textit{AdaMRA} on the \texttt{Byte-Level Text Classification} task and \texttt{Image Classification} task. $H$ denotes the number of heads. The best model is in boldface.}
   \vspace{10pt}
 \centering
  \begin{tabular}{c|c|c|c|c}
    \toprule
    ID&$H$& Compression Rate & Accu (Image) & Accu (Text) \\
    \midrule
    1&2& (\nicefrac{1}{4}, \nicefrac{1}{32}) &  45.3 & 66.8 \\ 
    2&3& (\nicefrac{1}{8}, \nicefrac{1}{16}, \nicefrac{1}{32}) &  45.5 &66.3 \\ 
    3&3& (\nicefrac{1}{2}, \nicefrac{1}{8}, \nicefrac{1}{32}) &  \textbf{46.0} & \textbf{68.4} \\ 
	4&4& (\nicefrac{1}{2}, \nicefrac{1}{4}, \nicefrac{1}{8}, \nicefrac{1}{16}) &  44.9 &67.8 \\ 
	5&4& (\nicefrac{1}{4}, \nicefrac{1}{8}, \nicefrac{1}{16}, \nicefrac{1}{32}) &  45.2 &68.2\\ 
	6&4& (\nicefrac{1}{8}, \nicefrac{1}{16}, \nicefrac{1}{32}, \nicefrac{1}{64}) &  44.0 &67.3\\ 
	7&6&(\nicefrac{1}{2}, \nicefrac{1}{4}, \nicefrac{1}{8}, \nicefrac{1}{16}, \nicefrac{1}{32}, \nicefrac{1}{64})&44.9&67.6\\  
	8&7&(\nicefrac{1}{2}, \nicefrac{1}{4}, \nicefrac{1}{8}, \nicefrac{1}{16}, \nicefrac{1}{32}, \nicefrac{1}{64}, \nicefrac{1}{128})&45.0&65.7\\ 
	\midrule
	9&3& (\nicefrac{1}{2}, \nicefrac{1}{2}, \nicefrac{1}{2}) &38.98&64.11 \\ 
	10&3& (\nicefrac{1}{4}, \nicefrac{1}{4}, \nicefrac{1}{4}) &42.38&63.38 \\ 
    \bottomrule
  \end{tabular} 
  \end{table} 

To understand the impact of compression rate $c$, we conduct ablation experiments on two tasks in the LRA benchmark, i.e., \texttt{Byte-Level Text Classification} and \texttt{Image Classification}.
We experiment with a various number of attention heads and vary the compression rate $c_h$ of each head. We use the same $c_h$ across all layers. 
As indicated by the results in Table 4, the choice of compression rate is crucial for the final performance. However, compared to vanilla Transformer, all configurations achieve consistent improvement on both tasks (see Table 1). 

Besides, there are few things to notice: 
i) Model 5 outperforms Model 4\&6, and Model 3 outperforms Model 2, indicating a benefit in using a moderate compression rate and using an extremely low compression rate cause a significant performance drop. We speculate that using an extremely low compression rate might lose too much information. 
ii) Model 3 outperforms Model 4-8, which means \textit{AdaMRA} does not perform better as the number of heads $H$ increases. This indicates that having multiple attention heads is effective, but a too large number of heads hurts.
iii) When we use a relatively low compression rate, the resulting model's (Model 3\&5) performance already outperforms all other Transformer models. This suggests that we can decrease the compression rate to a certain extent, which further increases the efficiency advantage of \textit{AdaMRA} over vanilla Transformer.  
iv) One can notice a significant accuracy drop when using the single-resolution (Model 9\&10), which indicates that multi-resolution attention head is beneficial for capturing hierarchically structured information.

\paragraph{Architecture Design} 
\begin{wraptable}{r}{5.5cm}
\vspace{-10pt}
\caption{Ablation study for architecture}
\centering
\begin{tabular}{ccc}
\toprule
Model&Image& Text\\
\midrule
Transformer&38.20&65.02\\
\midrule
\textit{Rand}&41.40&64.64\\
\textit{Softmax} &41.19&65.56\\
\textit{ELU+1} &43.73&66.33\\
\textit{ReLU}&\textbf{46.00}&\textbf{68.44}  \\
\bottomrule
\end{tabular}
\label{table:ta2}
\end{wraptable}

To understand the importance of each component, we conduct ablation experiments for the AdaMRA architecture. In Table 5, 
\textit{Rand} means randomly assigning each query to attention heads; \textit{Softmax} means adding multi-resolution approach into vanilla attention mechanism; \textit{ELU+1} \citep{katharopoulos2020transformers} means employing $elu+1$ as feature function $\phi$ in Eq. 7; \textit{ReLU} means using $ReLU$ as $\phi$. As indicated by the results, the privilege of AdaMRA comes from the learned routing and multi-resolution attention simultaneously. Using ReLU as the feature function is also advantageous, without which we only have small gains over the vanilla model. The multi-resolution method is also compatible with other attention mechanisms (e.g., vanilla attention) to a certain extent, and we leave it for future work.

\section{Conclusion}
Transformer models are notoriously slow to train and deploy in practice because of its quadratic time and space complexity with respect to the sequence length. 
In this paper, we propose a novel and efficient structure \textit{AdaMRA}. We see a benefit to this approach in the domain of text, image, mathematical expressions, etc., with the model outperforming existing architectures. 
In particular, we have shown that our model achieves state-of-the-art speed-memory-accuracy trade-off.
The main limitation of this work is the additional hyperparameters (number of Heads $H$ and the compression rate of each head $c_h$). However, we empirically show that multiple configurations work fairly well.
The proposed method opens several research directions towards integrating multi-resolution memory into Transformers. Besides, the scalability of \textit{AdaMRA} enables application in tasks that require working with large inputs, fast training speed, or low memory footprints.

\bibliographystyle{plainnat} 
\bibliography{neurips_2021.bib}

\begin{thebibliography}{35}
\providecommand{\natexlab}[1]{#1}
\providecommand{\url}[1]{\texttt{#1}}
\expandafter\ifx\csname urlstyle\endcsname\relax
  \providecommand{\doi}[1]{doi: #1}\else
  \providecommand{\doi}{doi: \begingroup \urlstyle{rm}\Url}\fi

\bibitem[Beltagy et~al.(2020)Beltagy, Peters, and Cohan]{beltagy2020longformer}
Iz~Beltagy, Matthew~E Peters, and Arman Cohan.
\newblock Longformer: The long-document transformer.
\newblock \emph{arXiv preprint arXiv:2004.05150}, 2020.

\bibitem[Child et~al.(2019)Child, Gray, Radford, and
  Sutskever]{child2019generating}
Rewon Child, Scott Gray, Alec Radford, and Ilya Sutskever.
\newblock Generating long sequences with sparse transformers.
\newblock \emph{arXiv preprint arXiv:1904.10509}, 2019.

\bibitem[Choromanski et~al.(2020)Choromanski, Likhosherstov, Dohan, Song, Gane,
  Sarlos, Hawkins, Davis, Mohiuddin, Kaiser, et~al.]{choromanski2020rethinking}
Krzysztof Choromanski, Valerii Likhosherstov, David Dohan, Xingyou Song,
  Andreea Gane, Tamas Sarlos, Peter Hawkins, Jared Davis, Afroz Mohiuddin,
  Lukasz Kaiser, et~al.
\newblock Rethinking attention with performers.
\newblock \emph{arXiv preprint arXiv:2009.14794}, 2020.

\bibitem[Correia et~al.(2019)Correia, Niculae, and
  Martins]{correia2019adaptively}
Gon{\c{c}}alo~M Correia, Vlad Niculae, and Andr{\'e}~FT Martins.
\newblock Adaptively sparse transformers.
\newblock \emph{arXiv preprint arXiv:1909.00015}, 2019.

\bibitem[Dong et~al.(2018)Dong, Xu, and Xu]{dong2018speech}
Linhao Dong, Shuang Xu, and Bo~Xu.
\newblock Speech-transformer: a no-recurrence sequence-to-sequence model for
  speech recognition.
\newblock In \emph{2018 IEEE International Conference on Acoustics, Speech and
  Signal Processing (ICASSP)}, pages 5884--5888. IEEE, 2018.

\bibitem[Guo et~al.(2019)Guo, Qiu, Liu, Shao, Xue, and Zhang]{guo2019star}
Qipeng Guo, Xipeng Qiu, Pengfei Liu, Yunfan Shao, Xiangyang Xue, and Zheng
  Zhang.
\newblock Star-transformer.
\newblock \emph{arXiv preprint arXiv:1902.09113}, 2019.

\bibitem[Howard and Ruder(2018)]{howard2018universal}
Jeremy Howard and Sebastian Ruder.
\newblock Universal language model fine-tuning for text classification.
\newblock \emph{arXiv preprint arXiv:1801.06146}, 2018.

\bibitem[Katharopoulos et~al.(2020)Katharopoulos, Vyas, Pappas, and
  Fleuret]{katharopoulos2020transformers}
Angelos Katharopoulos, Apoorv Vyas, Nikolaos Pappas, and Fran{\c{c}}ois
  Fleuret.
\newblock Transformers are rnns: Fast autoregressive transformers with linear
  attention.
\newblock In \emph{International Conference on Machine Learning}, pages
  5156--5165. PMLR, 2020.

\bibitem[Kitaev et~al.(2020)Kitaev, Kaiser, and Levskaya]{kitaev2020reformer}
Nikita Kitaev, {\L}ukasz Kaiser, and Anselm Levskaya.
\newblock Reformer: The efficient transformer.
\newblock \emph{arXiv preprint arXiv:2001.04451}, 2020.

\bibitem[Krizhevsky et~al.(2009)Krizhevsky, Hinton,
  et~al.]{krizhevsky2009learning}
Alex Krizhevsky, Geoffrey Hinton, et~al.
\newblock Learning multiple layers of features from tiny images.
\newblock 2009.

\bibitem[Linsley et~al.(2018)Linsley, Kim, Veerabadran, and
  Serre]{linsley2018learning}
Drew Linsley, Junkyung Kim, Vijay Veerabadran, and Thomas Serre.
\newblock Learning long-range spatial dependencies with horizontal
  gated-recurrent units.
\newblock \emph{arXiv preprint arXiv:1805.08315}, 2018.

\bibitem[Maas et~al.(2011)Maas, Daly, Pham, Huang, Ng, and
  Potts]{maas2011learning}
Andrew Maas, Raymond~E Daly, Peter~T Pham, Dan Huang, Andrew~Y Ng, and
  Christopher Potts.
\newblock Learning word vectors for sentiment analysis.
\newblock In \emph{Proceedings of the 49th annual meeting of the association
  for computational linguistics: Human language technologies}, pages 142--150,
  2011.

\bibitem[Nangia and Bowman(2018)]{nangia2018listops}
Nikita Nangia and Samuel~R Bowman.
\newblock Listops: A diagnostic dataset for latent tree learning.
\newblock \emph{arXiv preprint arXiv:1804.06028}, 2018.

\bibitem[Parmar et~al.(2018)Parmar, Vaswani, Uszkoreit, Kaiser, Shazeer, Ku,
  and Tran]{parmar2018image}
Niki Parmar, Ashish Vaswani, Jakob Uszkoreit, Lukasz Kaiser, Noam Shazeer,
  Alexander Ku, and Dustin Tran.
\newblock Image transformer.
\newblock In \emph{International Conference on Machine Learning}, pages
  4055--4064. PMLR, 2018.

\bibitem[Peng et~al.(2021)Peng, Pappas, Yogatama, Schwartz, Smith, and
  Kong]{peng2021random}
Hao Peng, Nikolaos Pappas, Dani Yogatama, Roy Schwartz, Noah~A Smith, and
  Lingpeng Kong.
\newblock Random feature attention.
\newblock \emph{arXiv preprint arXiv:2103.02143}, 2021.

\bibitem[Qiu et~al.(2019)Qiu, Ma, Levy, Yih, Wang, and Tang]{qiu2019blockwise}
Jiezhong Qiu, Hao Ma, Omer Levy, Scott Wen-tau Yih, Sinong Wang, and Jie Tang.
\newblock Blockwise self-attention for long document understanding.
\newblock \emph{arXiv preprint arXiv:1911.02972}, 2019.

\bibitem[Radev et~al.(2013)Radev, Muthukrishnan, Qazvinian, and
  Abu-Jbara]{radev2013acl}
Dragomir~R Radev, Pradeep Muthukrishnan, Vahed Qazvinian, and Amjad Abu-Jbara.
\newblock The acl anthology network corpus.
\newblock \emph{Language Resources and Evaluation}, 47\penalty0 (4):\penalty0
  919--944, 2013.

\bibitem[Rae et~al.(2019)Rae, Potapenko, Jayakumar, and
  Lillicrap]{rae2019compressive}
Jack~W Rae, Anna Potapenko, Siddhant~M Jayakumar, and Timothy~P Lillicrap.
\newblock Compressive transformers for long-range sequence modelling.
\newblock \emph{arXiv preprint arXiv:1911.05507}, 2019.

\bibitem[Rajpurkar et~al.(2016)Rajpurkar, Zhang, Lopyrev, and
  Liang]{rajpurkar2016squad}
Pranav Rajpurkar, Jian Zhang, Konstantin Lopyrev, and Percy Liang.
\newblock Squad: 100,000+ questions for machine comprehension of text.
\newblock \emph{arXiv preprint arXiv:1606.05250}, 2016.

\bibitem[Roy et~al.(2021)Roy, Saffar, Vaswani, and Grangier]{roy2021efficient}
Aurko Roy, Mohammad Saffar, Ashish Vaswani, and David Grangier.
\newblock Efficient content-based sparse attention with routing transformers.
\newblock \emph{Transactions of the Association for Computational Linguistics},
  9:\penalty0 53--68, 2021.

\bibitem[Shazeer et~al.(2017)Shazeer, Mirhoseini, Maziarz, Davis, Le, Hinton,
  and Dean]{shazeer2017outrageously}
Noam Shazeer, Azalia Mirhoseini, Krzysztof Maziarz, Andy Davis, Quoc Le,
  Geoffrey Hinton, and Jeff Dean.
\newblock Outrageously large neural networks: The sparsely-gated
  mixture-of-experts layer.
\newblock \emph{arXiv preprint arXiv:1701.06538}, 2017.

\bibitem[Sukhbaatar et~al.(2019)Sukhbaatar, Grave, Bojanowski, and
  Joulin]{sukhbaatar2019adaptive}
Sainbayar Sukhbaatar, Edouard Grave, Piotr Bojanowski, and Armand Joulin.
\newblock Adaptive attention span in transformers.
\newblock \emph{arXiv preprint arXiv:1905.07799}, 2019.

\bibitem[Tay et~al.(2020{\natexlab{a}})Tay, Bahri, Yang, Metzler, and
  Juan]{tay2020sparse}
Yi~Tay, Dara Bahri, Liu Yang, Donald Metzler, and Da-Cheng Juan.
\newblock Sparse sinkhorn attention.
\newblock In \emph{International Conference on Machine Learning}, pages
  9438--9447. PMLR, 2020{\natexlab{a}}.

\bibitem[Tay et~al.(2020{\natexlab{b}})Tay, Dehghani, Abnar, Shen, Bahri, Pham,
  Rao, Yang, Ruder, and Metzler]{tay2020long}
Yi~Tay, Mostafa Dehghani, Samira Abnar, Yikang Shen, Dara Bahri, Philip Pham,
  Jinfeng Rao, Liu Yang, Sebastian Ruder, and Donald Metzler.
\newblock Long range arena: A benchmark for efficient transformers.
\newblock \emph{arXiv preprint arXiv:2011.04006}, 2020{\natexlab{b}}.

\bibitem[Vaswani et~al.(2017)Vaswani, Shazeer, Parmar, Uszkoreit, Jones, Gomez,
  Kaiser, and Polosukhin]{vaswani2017attention}
Ashish Vaswani, Noam Shazeer, Niki Parmar, Jakob Uszkoreit, Llion Jones,
  Aidan~N Gomez, Lukasz Kaiser, and Illia Polosukhin.
\newblock Attention is all you need.
\newblock \emph{arXiv preprint arXiv:1706.03762}, 2017.

\bibitem[Voita et~al.(2019)Voita, Talbot, Moiseev, Sennrich, and
  Titov]{voita2019analyzing}
Elena Voita, David Talbot, Fedor Moiseev, Rico Sennrich, and Ivan Titov.
\newblock Analyzing multi-head self-attention: Specialized heads do the heavy
  lifting, the rest can be pruned.
\newblock \emph{arXiv preprint arXiv:1905.09418}, 2019.

\bibitem[Vyas et~al.(2020)Vyas, Katharopoulos, and Fleuret]{vyas2020fast}
Apoorv Vyas, Angelos Katharopoulos, and Fran{\c{c}}ois Fleuret.
\newblock Fast transformers with clustered attention.
\newblock \emph{Advances in Neural Information Processing Systems}, 33, 2020.

\bibitem[Wang et~al.(2020{\natexlab{a}})Wang, Ye, Zhang, Zhang, and
  Smola]{wang2020transformer}
Chenguang Wang, Zihao Ye, Aston Zhang, Zheng Zhang, and Alexander~J Smola.
\newblock Transformer on a diet.
\newblock \emph{arXiv preprint arXiv:2002.06170}, 2020{\natexlab{a}}.

\bibitem[Wang et~al.(2020{\natexlab{b}})Wang, Li, Khabsa, Fang, and
  Ma]{wang2020linformer}
Sinong Wang, Belinda Li, Madian Khabsa, Han Fang, and Hao Ma.
\newblock Linformer: Self-attention with linear complexity.
\newblock \emph{arXiv preprint arXiv:2006.04768}, 2020{\natexlab{b}}.

\bibitem[Williams et~al.(2017)Williams, Nangia, and Bowman]{williams2017broad}
Adina Williams, Nikita Nangia, and Samuel~R Bowman.
\newblock A broad-coverage challenge corpus for sentence understanding through
  inference.
\newblock \emph{arXiv preprint arXiv:1704.05426}, 2017.

\bibitem[Xiong et~al.(2021)Xiong, Zeng, Chakraborty, Tan, Fung, Li, and
  Singh]{xiong2021nystr}
Yunyang Xiong, Zhanpeng Zeng, Rudrasis Chakraborty, Mingxing Tan, Glenn Fung,
  Yin Li, and Vikas Singh.
\newblock Nystr$\backslash$" omformer: A nystr$\backslash$" om-based algorithm
  for approximating self-attention.
\newblock \emph{arXiv preprint arXiv:2102.03902}, 2021.

\bibitem[Xu et~al.(2015)Xu, Ba, Kiros, Cho, Courville, Salakhudinov, Zemel, and
  Bengio]{xu2015show}
Kelvin Xu, Jimmy Ba, Ryan Kiros, Kyunghyun Cho, Aaron Courville, Ruslan
  Salakhudinov, Rich Zemel, and Yoshua Bengio.
\newblock Show, attend and tell: Neural image caption generation with visual
  attention.
\newblock In \emph{International conference on machine learning}, pages
  2048--2057. PMLR, 2015.

\bibitem[Ye et~al.(2019)Ye, Guo, Gan, Qiu, and Zhang]{ye2019bp}
Zihao Ye, Qipeng Guo, Quan Gan, Xipeng Qiu, and Zheng Zhang.
\newblock Bp-transformer: Modelling long-range context via binary partitioning.
\newblock \emph{arXiv preprint arXiv:1911.04070}, 2019.

\bibitem[Zaheer et~al.(2020)Zaheer, Guruganesh, Dubey, Ainslie, Alberti,
  Ontanon, Pham, Ravula, Wang, Yang, et~al.]{zaheer2020big}
Manzil Zaheer, Guru Guruganesh, Avinava Dubey, Joshua Ainslie, Chris Alberti,
  Santiago Ontanon, Philip Pham, Anirudh Ravula, Qifan Wang, Li~Yang, et~al.
\newblock Big bird: Transformers for longer sequences.
\newblock \emph{arXiv preprint arXiv:2007.14062}, 2020.

\bibitem[Zhao et~al.(2019)Zhao, Lin, Zhang, Ren, Su, and Sun]{zhao2019explicit}
Guangxiang Zhao, Junyang Lin, Zhiyuan Zhang, Xuancheng Ren, Qi~Su, and Xu~Sun.
\newblock Explicit sparse transformer: Concentrated attention through explicit
  selection.
\newblock \emph{arXiv preprint arXiv:1912.11637}, 2019.

\end{thebibliography}

\end{document}